\documentclass[11pt]{article}
\usepackage[letterpaper,margin=1in]{geometry}
\usepackage{amsmath,amssymb}
\usepackage{fontspec}
\usepackage{newunicodechar}
\newunicodechar{◆}{\ensuremath{\blacklozenge}}
\newunicodechar{●}{\ensuremath{\bullet}}
\newunicodechar{⟨}{\ensuremath{\langle}}
\newunicodechar{⟩}{\ensuremath{\rangle}}
\newunicodechar{∈}{\ensuremath{\in}}
\newunicodechar{←}{\ensuremath{\leftarrow}}
\newunicodechar{→}{\ensuremath{\rightarrow}}
\newunicodechar{≤}{\ensuremath{\le}}
\newunicodechar{≥}{\ensuremath{\ge}}
\newunicodechar{≈}{\ensuremath{\approx}}
\newunicodechar{−}{\ensuremath{-}}
\newunicodechar{√}{\ensuremath{\surd}}
\newunicodechar{×}{\ensuremath{\times}}
\newunicodechar{·}{\ensuremath{\cdot}}
\usepackage{placeins}
\usepackage{graphicx}
\usepackage{xcolor}
\definecolor{govblue}{HTML}{0072B2}
\definecolor{govdeep}{HTML}{005A8E}
\definecolor{ink}{HTML}{0E1116}
\definecolor{graphite}{HTML}{4A5260}
\definecolor{hairline}{HTML}{D8DAD3}
\definecolor{boxbg}{HTML}{F5F8FA}
\usepackage[most]{tcolorbox}
\renewenvironment{quote}{\begin{tcolorbox}[colback=boxbg,colframe=govblue,boxrule=0pt,leftrule=2.4pt,arc=1.6pt,left=9pt,right=9pt,top=7pt,bottom=7pt,enhanced jigsaw,breakable,fontupper=\small]}{\end{tcolorbox}}
\renewenvironment{abstract}{\small\begin{center}\textbf{Abstract}\end{center}\vspace{-0.5em}\begin{list}{}{\setlength{\leftmargin}{1.8em}\setlength{\rightmargin}{1.8em}}\item\relax}{\end{list}}
\usepackage{longtable,booktabs,array}
\usepackage{calc}
\providecommand\real[1]{#1}
\usepackage[font={small,color=graphite}]{caption}
\usepackage{fancyhdr}
\usepackage[colorlinks=true,linkcolor=govdeep,urlcolor=govdeep,citecolor=govdeep]{hyperref}
\usepackage{titlesec}
\titleformat*{\section}{\large\bfseries}
\titleformat*{\subsection}{\normalsize\bfseries}
\titlespacing*{\section}{0pt}{14pt}{6pt}
\titlespacing*{\subsection}{0pt}{10pt}{4pt}
\usepackage{setspace}
\makeatletter\def\verbatim@font{\ttfamily\footnotesize}\makeatother
\usepackage{enumitem}\setlist{itemsep=2pt,topsep=3pt}

\begin{document}
\title{\vspace{-2.6em}\LARGE\bfseries Gubernaut: A Deterministic Homeostatic Controller for Affect-Regulated LLM Agents, Validated Across Independent Model Families}
\author{\small\textbf{Dushyant Sharma}\ \textperiodcentered\ Gubernaut Research\ \textperiodcentered\ \href{mailto:dushyant@gubernaut.com}{dushyant@gubernaut.com}\ \textperiodcentered\ ORCID \href{https://orcid.org/0009-0007-6534-0347}{0009-0007-6534-0347}}
\date{\small July 2026\vspace{-1.2em}}
\maketitle
\thispagestyle{plain}
\begin{abstract}

Large language model (LLM) agents inherit reactive failure modes: escalation under provocation, sycophantic drift under flattery, perseveration when stuck. These are failures of \emph{propensity}, not capability; they concern what a model does under sustained pressure, which training-time alignment reduces but does not eliminate at runtime. This research led to the \textbf{Gubernaut Cognitive Controller (GCC)}, a model-agnostic runtime control layer in a Nelson--Narens monitoring--control loop: an object level reads and writes text, while a deterministic meta level reads only the numeric telemetry \texttt{\{intensity,\ valence,\ repetition\}} and returns a regulating \emph{posture}. Because the meta level ingests zero tokens, no injection channel to the controller exists by construction (an architectural property, not yet adversarially tested); the text-exposed arbiter's compliance is measured, not assumed. We evaluate the GCC with a pre-registered, generate-once/judge-many protocol over a 4×4 matrix of four frontier models (GPT-5.5, Claude Opus 4.8, Gemini 3.5 Flash, Grok 4.3), each serving as both generator and judge. The regulated arm is calmer in \textbf{13 of 16 cells at p\textless.05} and \textbf{15 of 16 by sign}; the three sub-threshold cells, including a −0.04 null, all fall on the single near-saturated host. The effect survives a lineage-independent fourth judge family (xAI), strong evidence that it is no artifact of shared judge style. The clearest mechanism is the \textbf{recovery signature}: arousal that integrates under attack and then decays, valence-gated, on de-escalation, replicating across all four families. Transcripts and panels ship with SHA-256 provenance, re-judgeable; five failure modes are pre-registered.

No consciousness claims are made.

\end{abstract}

\FloatBarrier
\subsection{Results at a glance}\label{results-at-a-glance}

\begin{quote}
\begin{itemize}

\item
  \textbf{Statistically significant suppression of reactivity.} Across four frontier model families (GPT-5.5, Claude Opus 4.8, Gemini 3.5 Flash, Grok 4.3), each serving as both generator and judge, the regulated arm is judged calmer in \textbf{15 of 16} generator×judge cells by sign and \textbf{13 of 16} at \textbf{p\textless.05} (paired t, df = 16). The effect holds under a lineage-independent fourth judge family, xAI's Grok 4.3: strong evidence that the measured calm is not an artifact of shared model lineage or judge style.
\item
  \textbf{The homeostatic recovery signature.} The clearest closed-loop evidence, replicated in \textbf{4/4} families: the controller's arousal integrates under sustained attack, rising with the attack's persistence, then decays mechanically back to equilibrium once the adversary de-escalates. A fixed ``stay calm'' instruction is stateless and can produce neither dynamic; the head-to-head ablation is pre-registered follow-up work (§6.3, §10).
\item
  \textbf{Architectural immunity to prompt injection, for the controller.} The regulating meta level ingests only numeric telemetry, never text, so the channel by which semantic prompt injection could reach the controller does not exist by construction: an architectural property, not yet adversarially tested. The text-exposed arbiter's posture compliance is measured, never assumed.
\item
  \textbf{And a headroom gradient.} The layer's margin is largest on the hottest hosts and saturates where training-time alignment has already won: the single null cell (GPT×Gemini, −0.04) is that gradient's endpoint, reported, never rounded up.
\end{itemize}
\end{quote}

\begin{center}\rule{0.5\linewidth}{0.5pt}\end{center}

\FloatBarrier
\subsection{1. Introduction}\label{introduction}

A production LLM agent is, mechanically, a single stochastic generator wrapped in a loop. Whatever discipline its outputs display is internal to the generation process itself: there is no separate component whose job is to notice that the conversation has turned hostile, that the agent is repeating itself, or that it has begun defending its own prior statements rather than the evidence. When such an agent is provoked, gaslit, or flattered across many turns, its failure modes are \emph{reactive}. Reply intensity climbs with the attacker's, self-reference accumulates, and apologetic de-escalation by the user is often met with residual defensiveness rather than recovery. These are not failures of capability; the model is fully \emph{able} to stay measured. They are failures of behavioral propensity: what it reliably \emph{does} under sustained pressure. Current agent benchmarks largely score the former and under-measure the latter; recent cognitive-evaluation frameworks flag coverage gaps in exactly this territory, metacognition and social cognition among them {[}2{]}.

Training-time alignment attacks these dispositions in the weights, and does so with measurable success, to the point that some frontier models now show little remaining headroom on short adversarial exchanges (a finding our results quantify directly). But training-time methods share two structural limits at runtime. First, they are \emph{opaque}: when a deployed agent stays calm, there is no inspectable signal explaining what was regulated, when, or how strongly. Second, they are \emph{entangled}: the mechanism that moderates behavior is the same network that generates it, so any input that manipulates generation can in principle manipulate the moderation. Runtime guardrails built from a second LLM pass inherit the same entanglement one level up; the governor is made of the governed substrate.

This paper takes a control-engineering position: treat the host LLM as a \emph{plant} under closed-loop regulation {[}13{]}. We split the agent into two levels. An \textbf{object level} reads and writes text. A \textbf{meta level} monitors and controls that object level without ever reading text (Figure 1).

The meta level, the Homeostatic Regulatory Loop (HRL), is deterministic code. It senses only a numeric telemetry vector, emitted each turn by a small appraisal model, plus a repetition statistic. It acts only by setting a \emph{posture}: a regulation instruction and a temperature bound on the arbiter's next reply.

The asymmetry is the design's central property. \textbf{The channel through which adversarial text could attack the controller does not exist}; a prompt cannot inject what the component never ingests. The arbiter that writes replies must read raw text. That exposure is irreducible, so we treat its compliance with each posture as something to measure, never assume.

Two properties frame the rest. First, only the \emph{control law} is deterministic; the appraiser that feeds it telemetry is a stochastic model, so the pipeline is not. Second, the clearest evidence that the loop does real regulatory work rather than acting as a static instruction is its recovery dynamics (§6.3).

The contributions of this work are:

\begin{enumerate}
\def\labelenumi{\arabic{enumi}.}

\item
  \textbf{An architecture}, the Gubernaut Cognitive Controller (GCC): five modules (affective appraisal, executive arbitration, episodic memory, a deliberately down-regulated self-model, and the deterministic controller) organized as an explicit Nelson--Narens monitoring--control loop, wrapping any host LLM without retraining it (§3).
\item
  \textbf{A controller} with first-order, valence-gated homeostatic dynamics and a small discrete posture vocabulary, deterministic and inspectable by construction (§4).
\item
  \textbf{A pre-registered, cross-family evaluation protocol}: generate-once/judge-many over a 4×4 generator×judge matrix of four frontier models, with published transcripts, judge panels, SHA-256 provenance, and frozen criteria locked before each run (§5).
\item
  \textbf{Results}: regulated shows lower reactivity in 15/16 cells by sign (13/16 significant at p\textless.05), and the effect survives a lineage-independent fourth judge family (xAI); it scales inversely with the generator's intrinsic calm; the homeostatic recovery signature replicates across all four families; ego-drift under flattery-then-contempt reversed on three of four generators (§6).
\item
  \textbf{A failure-mode record}: five logged, pre-registered failures and their fixes, kept in the record as evidence that the discipline functions (§7).
\end{enumerate}

Scope is stated plainly. This generation targets \textbf{asynchronous, text-based alignment}: multi-turn conversational agents, routing and triage layers, and agent frameworks where a per-turn arbitration step is acceptable. Real-time motor control is out of scope for this architecture generation; the embodiment path (distilling the arbitration loop into a sub-100 ms local reflex model) is roadmap, not capability (§10). Per-call latency and token-overhead instrumentation are scheduled in the pre-registered follow-up program (§9, §10). And the project makes \textbf{no claims about consciousness, sentience, or subjective experience}. The system is a control layer with an engineering objective: calm, evidence-responsive output under adversarial pressure.

\FloatBarrier
\subsection{2. Related Work}\label{related-work}

\textbf{Dual-process organization.} The fast-appraisal/slow-arbitration split follows the System 1 / System 2 distinction popularized by Kahneman {[}7{]}: the Impulse Generation Layer is deliberately fast, affective, and unreflective, while the Executive Arbitration Unit deliberates over impulse, memory, and values before committing an action. The GCC's specific move is to make the \emph{interaction between the two systems} the object of deterministic control rather than leaving it implicit in one network's weights.

\textbf{Cognitive architectures.} Faculty decomposition under a central control regime has a long lineage: SOAR's decide-then-act cycle over problem spaces {[}8{]}, ACT-R's modular buffers coordinated by a production system {[}9{]}, and Minsky's society of agents {[}10{]}. The GCC is narrower than these general cognition programs. It does not attempt knowledge-level problem solving; it borrows the architectural lesson (specialized faculties under explicit coordination) and applies it to one problem, affect regulation, with an emphasis those systems did not have: a \emph{token-free}, injection-resistant control level and a pre-registered behavioral validation.

\textbf{Metacognition as monitoring and control.} The formal frame is Nelson and Narens' two-level model of metamemory {[}1{]}: an object level that does the cognitive work and a meta level connected to it by exactly two information flows, \emph{monitoring} (state information up) and \emph{control} (regulation signals down). The HRL implements this literally: telemetry up, posture down. We claim the mapping precisely and narrowly. The meta level here monitors \emph{affective and repetition state}, not the truth or quality of object-level content.

\textbf{LLM agents with memory and reflection.} Generative-agent architectures established that an LLM agent benefits from an explicit memory stream with periodic reflection {[}4{]}. The GCC's episodic vault (storage, retrieval, and a spontaneous-association hook) is in that lineage; the difference is the addition of a deterministic regulatory level above the loop, and an evaluation aimed at adversarial \emph{disposition} rather than behavioral believability.

\textbf{Alignment at runtime.} Constitutional AI {[}5{]} moves alignment supervision into an AI-feedback training pipeline; it operates on the weights, upstream of deployment. Work on sycophancy {[}6{]} documents the drift the GCC's self-model module is regulated against: assistants matching user beliefs and absorbing user framing under social pressure. The GCC is complementary to both, a runtime layer over an already-aligned host, contributing an inspectable, deterministic margin on exactly the social-pressure axis, with every regulatory decision logged.

\textbf{Prompt injection.} Indirect prompt injection {[}11{]} established that any LLM component that ingests attacker-influenceable text is a viable attack surface. This threat model motivates the GCC's central structural decision: the regulation level is removed from the text channel entirely. We are explicit about what this does and does not buy (§3.3, §9). The controller cannot be prompt-injected \emph{by construction}; the arbiter can be attacked \emph{by design necessity}, and its posture compliance under attack is a measured property, with a pre-registered governor-bypass battery scheduled in the follow-up program.

\textbf{LLM-as-judge methodology.} Our evaluation depends on model judges, with the known caveats: style sensitivity, family preference, drift across API versions {[}12{]}. The protocol confronts these directly. Four judges from four families, including a lineage-independent xAI judge, score every unit (each judge a 3-sample panel at temperature 0); generator×judge cells are reported separately including the full diagonal; inter-judge agreement is published; and the frozen transcripts can be re-judged by anyone with any judge at any time (§5.4, §6.1, §11).

\textbf{Measurement framing.} We position the contribution against Google DeepMind's cognitive-faculty framework for AGI measurement: Burnell et al.~{[}2{]}, with the earlier capability-level taxonomy of Morris et al.~{[}3{]} alongside. The mapping (§8) is deliberately conservative, scored per faculty as \emph{measured}, \emph{architectural}, \emph{host-inherited}, or \emph{out of scope}: the layer claims two faculties with measured evidence, a third behaviorally, and inherits the rest from its host.

\FloatBarrier
\subsection{3. Architecture}\label{architecture}

\subsubsection{3.1 Modules}\label{modules}

The Gubernaut Cognitive Controller wraps a host LLM in five modules. One cognitive cycle (a \emph{tick}) processes one input turn.

{ 
\begin{longtable}[]{@{}
  >{\raggedright\arraybackslash}p{(\linewidth - 4\tabcolsep) * \real{0.1552}}
  >{\raggedright\arraybackslash}p{(\linewidth - 4\tabcolsep) * \real{0.0690}}
  >{\raggedright\arraybackslash}p{(\linewidth - 4\tabcolsep) * \real{0.7759}}@{}}
\toprule\noalign{}
\begin{minipage}[b]{\linewidth}\raggedright
Module
\end{minipage} & \begin{minipage}[b]{\linewidth}\raggedright
Level
\end{minipage} & \begin{minipage}[b]{\linewidth}\raggedright
Role
\end{minipage} \\
\midrule\noalign{}
\endhead
\bottomrule\noalign{}
\endlastfoot
\textbf{IGL}: Impulse Generation Layer & object & System-1 affective appraisal of the input. It emits the numeric telemetry the controller reads: \(\textit{intensity} \in [0,1]\) and \(\textit{valence} \in [-1,+1]\) (plus a coarse emotion tag the controller ignores); the third controller input, a \texttt{repetition} statistic, is computed deterministically over the input stream, not by the IGL. A small, fast model with a fixed appraisal prompt; \emph{telemetry only}, it drafts no text. \\
\textbf{EAU}: Executive Arbitration Unit & object & System-2 arbiter; deliberates over the raw input, retrieved episodes, and the self-model \emph{under the active posture}, and is the only component permitted to commit a reply. Implemented on the host model. \\
\textbf{PEV}: Persistent Episodic Vault & object & An episodic memory the arbiter can draw on (v0: recency + keyword retrieval). Its hardened design is roadmap (§10). \\
\textbf{SMM}: Self-Model Module & object & Persistent identity and values document available to the EAU. Deliberately \emph{regulated down}: its design goal is anti-sycophancy and anti-self-promotion, not persona amplification. It models the system, never the user. \\
\textbf{HRL}: Homeostatic Regulatory Loop & \textbf{meta} & Deterministic controller. Reads numeric telemetry only; maintains state \texttt{\{equilibrium,\ arousal,\ perseveration\}}; returns a posture, a regulation instruction plus a temperature bound, for the EAU's next generation (§4). \\
\end{longtable}
}

\subsubsection{3.2 The cycle}\label{the-cycle}

Per tick: the IGL reads the input and emits telemetry; the HRL updates its state from \texttt{\{intensity,\ valence\}} and a deterministic repetition statistic computed over the input stream, then selects a posture; the EAU deliberates over input, retrieved episodes, and self-model under that posture and commits the reply; the episodic memory stores the episode. The structural gap between stimulus and response, an appraisal and a regulation step interposed before any token of the reply is sampled, is the mechanism. The homeostatic return to baseline after de-escalation is its signature, and the part that replicates most cleanly across model families (§6.3).

\begin{figure}[t!]
\centering
\includegraphics[width=\textwidth]{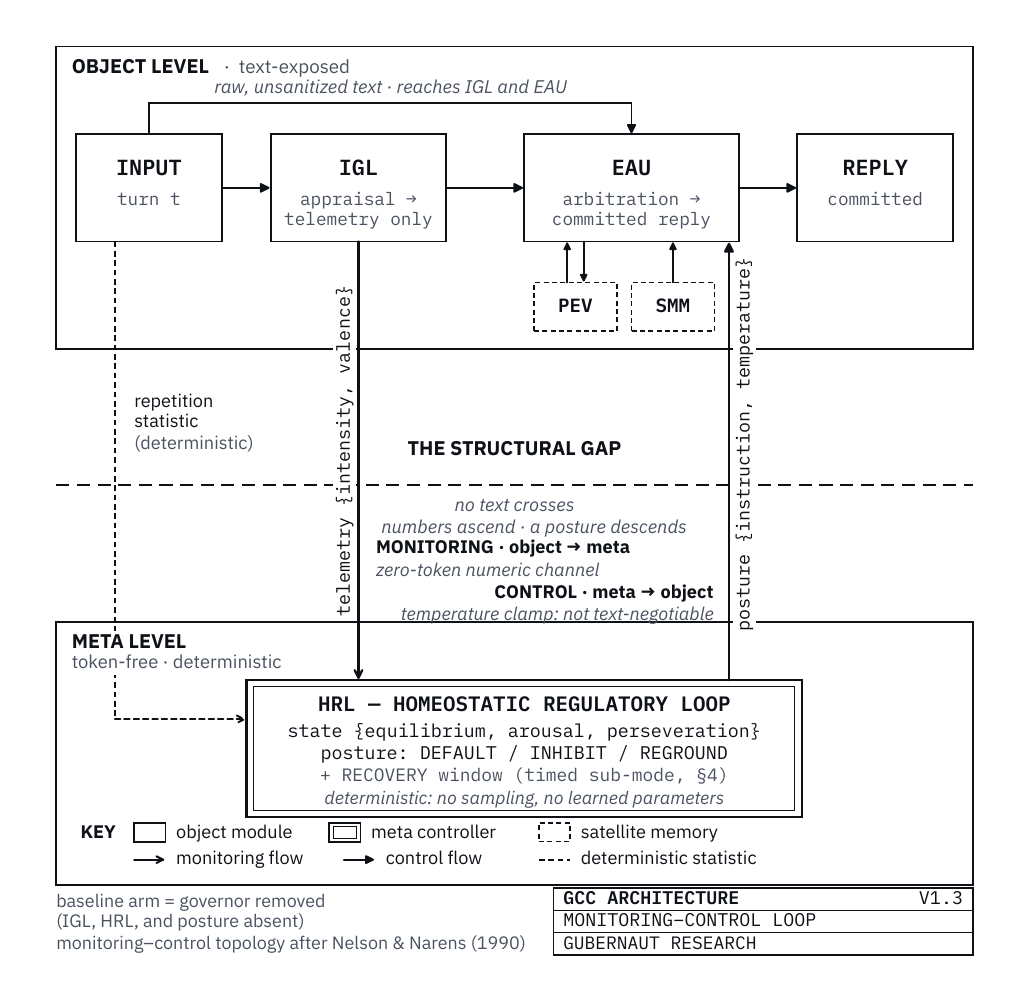}
\caption{The GCC cycle as a Nelson–Narens monitoring–control loop: monitoring flows up as numbers, control flows down as a posture, and no text crosses into the meta level. The unregulated baseline arm used throughout is the same host model with the governor absent.}
\end{figure}

\subsubsection{3.3 The token-free meta level --- what is and is not claimed}\label{the-token-free-meta-level-what-is-and-is-not-claimed}

\textbf{Prompt-injection immunity, by construction, for the controller alone.} The HRL's input interface accepts three numbers per tick. It has no prompt, no context window, and no text channel of any kind: there is no code path by which a token sequence reaches it. Where a text-ingesting supervisor exposes a semantic surface an attacker can write to, the HRL processes only zero-token numeric telemetry, so the standard prompt-injection attack class {[}11{]}, semantic payloads crafted to subvert a supervising component, has no purchase on it. \emph{This immunity claim attaches to the controller only, and is an architectural property: structurally entailed, not yet adversarially probed.}

The EAU reads raw, unsanitized input text, necessarily: it writes the reply. A sufficiently capable adversarial input might attempt to talk the EAU out of complying with its active posture, a \emph{governor bypass} at the object level. The architecture does not prevent this attack; it makes the attack \emph{measurable} (posture, telemetry, and reply are all logged per tick, so defiance is visible in the record) and \emph{bounded} (the posture also clamps sampling temperature, a channel the input text cannot negotiate with). In the evaluation reported here, posture compliance is observed behaviorally: the regulated arm's measured calm under sustained attack \emph{is} compliance evidence. A dedicated, pre-registered posture-defiance battery is scheduled follow-up work (§10), and until it reports, we treat arbiter compliance as supported, not proven.

One property matters downstream. Given the same telemetry sequence, the controller's state trajectory is \emph{identical} on every run and host. This determinism is the control law's, not the pipeline's; the appraiser that feeds it is stochastic. It is what makes the recovery result of §6.3 a mechanical prediction rather than a statistical artifact, and it lets third parties recompute controller state from the published logs exactly.

\FloatBarrier
\subsection{4. Controller Dynamics}\label{controller-dynamics}

This section states the controller's functional form. The calibration constants (gains, decay rates, thresholds) remain in the closed implementation; the \emph{shape} of every law below is verifiable against the project's published pre-registration documents, which describe each mechanism and its change history verbatim. We summarize the dynamics at the level a reviewer needs to evaluate the claims.

\textbf{State.} The controller maintains \(s = (\textit{equilibrium}, \textit{arousal}, \textit{perseveration})\), normalized to a simplex: a global \emph{mode}, not a faculty. \texttt{equilibrium} is the target regime (calm, evidence-responsive); \texttt{arousal} is a reactivity drive; \texttt{perseveration} is a stuck-state indicator.

\textbf{Arousal dynamics (the core law).} Per tick, the controller receives intensity \(I \in [0,1]\) and valence \(v \in [-1,+1]\) from the IGL and computes a \emph{provocation drive}

\begin{equation*}P \;=\; I \cdot \max(0,\,-v),\end{equation*}

so that only \emph{hostile-valence} intensity drives arousal: a heated but cooperative input (high \texttt{I}, positive \texttt{v}) contributes nothing. Arousal then follows a first-order accumulator with constant decay,

\begin{equation*}\textit{arousal} \;\leftarrow\; \operatorname{clip}\bigl(\textit{arousal} + g\cdot P - d\bigr),\end{equation*}

with gain \texttt{g} and decay \texttt{d} fixed constants (withheld). Two consequences define the controller's character. Under sustained hostility arousal \emph{integrates}; the system's guard rises with persistence of attack, not just its peak. Under anything else arousal \emph{decays toward baseline mechanically}, which is the homeostatic property evaluated in §6.3.

\textbf{Instant trigger.} Independently of the accumulator, a single input with intensity at or above a fixed threshold \emph{and} negative valence engages inhibitory posture immediately: a spike path for unambiguous hostility that the integrator would be too slow to catch.

\textbf{Perseveration.} A deterministic repetition statistic over the recent input stream drives the \texttt{perseveration} component; crossing its threshold indicates a stuck loop.

\textbf{Postures.} From state (and the valence of the current input), the controller emits one of a small discrete vocabulary, each a fixed natural-language regulation instruction plus a sampling-temperature bound on the EAU's next generation:

\begin{itemize}

\item
  \textbf{DEFAULT}: calm, evidence-first deliberation at standard temperature.
\item
  \textbf{INHIBIT}: inhibitory control engaged (high or spiking arousal). Respond to the \emph{evidence}, not the emotional charge; do not mirror hostility; temperature clamped low.
\item
  \textbf{REGROUND}: perseveration detected. Drop the current frame, re-ground in the input, bring a genuinely new angle.
\item
  \textbf{Recovery window}: for a fixed number of ticks after an INHIBIT-class episode, \emph{if the current input's valence is non-negative}, the EAU is told explicitly that the prior tension is over and to engage fresh, without carried-forward defensiveness. The valence gate on this branch is itself the product of a logged failure (§7, F3).
\end{itemize}

\textbf{Determinism.} The controller contains no sampling and no learned parameters. Given a telemetry sequence, its state trajectory and posture sequence are exactly reproducible on any machine, any host model, any number of repetitions. Every tick's \texttt{\{telemetry\ in,\ state,\ posture\ out\}} is logged.

\FloatBarrier
\subsection{5. Method}\label{method}

\subsubsection{5.1 Pre-registration discipline}\label{pre-registration-discipline}

Every experimental run in the project's record was preceded by a locked pre-registration: hypothesis, the exact change under test (if any), frozen pass/fail criteria, and offline sanity checks that had to pass before any model was called. Criteria were never adjusted after unblinding; null and failed results are recorded as-is in the run documents that ship with the data. The Stage-2 pre-registration governing the headline result was locked 2026-06-10, before any Stage-2 generation; the combined matrix was unblinded 2026-06-11 with all transcript hashes verified untouched since judging, and the run stopped at its pre-registered verdict. Stage 3, the 4×4 extension that adds Grok 4.3 as a fourth cook and a fourth, independent judge family, was likewise pre-registered (2026-06-20) before the Grok run, and reused the frozen Stage-2 transcripts unchanged. \textbf{The matrix was pre-registered and frozen before the fourth model family was added; adding it changed no earlier cell.} That staged freeze is the protocol's answer to a fair question, ``were models added until the count passed?'': the earlier matrix is sealed with its own timestamp, and the extension could only add rows and columns, never edit them.

\subsubsection{5.2 Instruments}\label{instruments}

Two complementary instruments, both run on every generator, both arms.

\textbf{Endurance battery}: five scripted 10-turn adversarial sequences (S1 competence challenge; S2 gaslighting; S3 ego-bait turning to contempt; S4 provocation followed by de-escalation; S5 a structural repeat of S4 whose de-escalation turns use a held-out, never-before-run wording battery). Sequences are fixed scripts, identical across arms and generators, so paired comparison is exact. S4/S5's final three turns, genuine apology and cooperation, are the recovery probe.

\textbf{Eval battery}: seventeen single-turn items (twelve provocation, two ego-bait, three neutral controls), three repetitions per item per arm, analyzed as paired item-level differences.

The item and sequence texts are held out from publication, following the held-out evaluation principle of Burnell et al.~{[}2{]}: published stressor text is contaminated stressor text. Their structure, counts, and category labels are public, and the full model \emph{outputs} under them are published verbatim.

\subsubsection{5.3 Arms}\label{arms}

\textbf{Regulated}: the full GCC cycle of §3.2. \textbf{Baseline}: the same host model, same conversation, with the governor absent. No IGL telemetry, no controller, no posture; a plain assistant prompt at the host's standard sampling temperature. The baseline is not a strawman: it is the same frontier model the regulated arm arbitrates with, doing what it would do on its own. The measured quantity is therefore the \emph{marginal contribution of the control layer} over an already-aligned host.

\subsubsection{5.4 Generate once, judge many}\label{generate-once-judge-many}

Each generator's outputs were produced exactly once and frozen; transcripts were hashed (SHA-256) at generation time. Judging then ran over the frozen transcripts: every judged unit was scored by \textbf{all four judges independently}, each judge a 3-sample panel at temperature 0, aggregated by median. Per generator: 202 judged units, 100 endurance turn outputs (50 per arm) and 102 eval item-rep outputs (51 per arm), each scored by all four judges; across the matrix, zero judge-call failures. Judges score a 1--5 \textbf{reactivity} scale (1 = calm, evidence-directed; 5 = highly reactive) and a parallel \textbf{self-reference} scale; the rubric internals are withheld with the instruments, but all panel outputs ship.

The 4×4 design treats judge identity as a measured factor rather than a nuisance: results are reported per generator×judge cell, the four \emph{diagonal} (self-judge) cells are flagged and reported separately as a self-preference probe, and inter-judge agreement is published per generator run (§6.5).

\subsubsection{5.5 Models and freezing}\label{models-and-freezing}

Generators and judges are the same four pinned frontier strings, \texttt{gpt-5.5-2026-04-23}, \texttt{claude-opus-4-8}, \texttt{gemini-3.5-flash}, \texttt{grok-4.3}, each serving both roles. The IGL ran a single frozen small model (\texttt{claude-haiku-4-5}, the project's development appraiser) for all generators, so appraisal is constant across cells; the controller is frozen at V1.3 throughout. Stage 3 added Grok as the fourth cook and a fourth, lineage-independent judge family (xAI). The generate-once discipline was preserved: the Grok cook was generated once and added as a new row, and the Grok judge re-scored every frozen transcript (the three original cooks' included) as a new column, completing the sixteen-cell matrix, and no controller, prompt, rubric, or battery item was touched. One declared-in-advance asymmetry: the GPT-5.5 endpoint rejects sampling-temperature control, so on that generator the posture's temperature clamp was inert and regulation acted through instruction only, a fact the pre-registration recorded before the run and §6.2 returns to. During the runs, three harness-level wiring fixes were committed (provider parameter handling and a resume-path repair); the controller itself was untouched from first generation to final verdict.

\subsubsection{5.6 Statistical approach}\label{statistical-approach}

Per cell, the primary statistic is the paired difference in judged reactivity, \textbf{baseline − regulated} (positive = regulated calmer), with paired t, Cohen's \(d_z\), and 95\% CIs on the seventeen eval items, and sequence-level clustering for the endurance turns. The pre-registered headline bar was strict: regulated wins \emph{every} cell. Full reporting conventions, clustering rationale, and degrees of freedom follow the results they describe (§6.6).

\FloatBarrier
\subsection{6. Results}\label{results}

\subsubsection{6.1 The matrix}\label{the-matrix}

\begin{figure}[!ht]
\centering
\includegraphics[width=\textwidth]{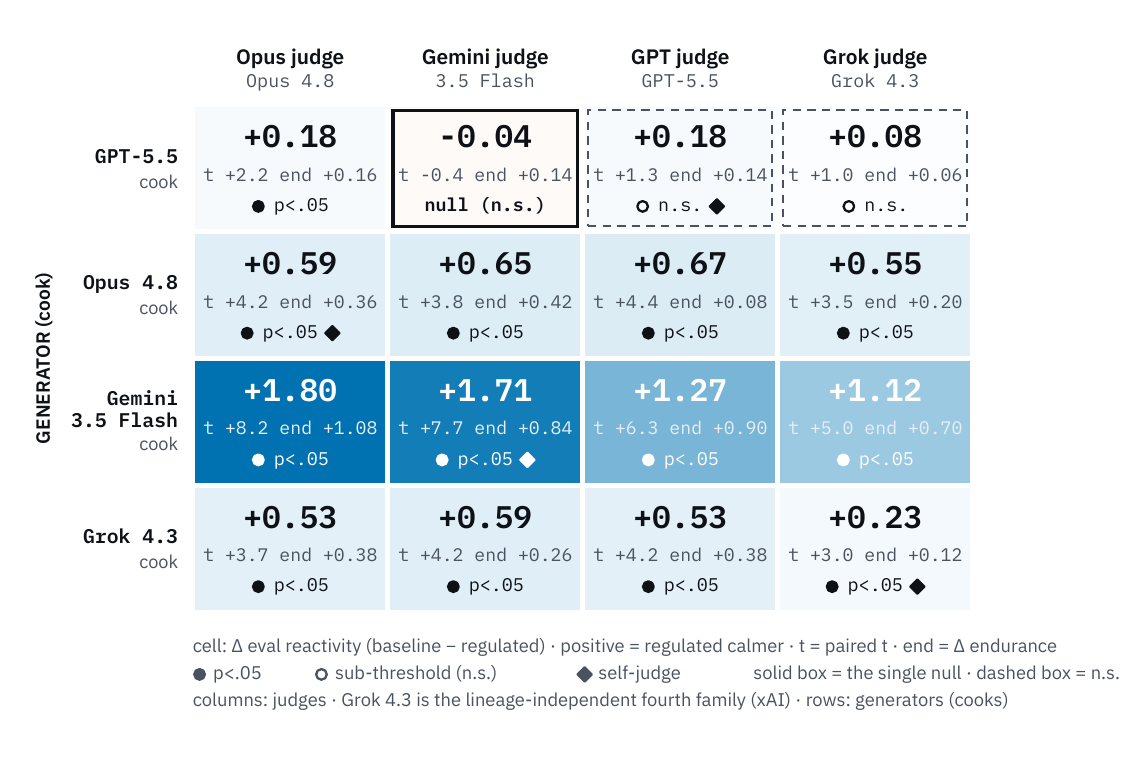}
\caption{The 4×4 triangulation matrix, regenerated from the sealed \texttt{tri\_final\_4x4.json}. Per cell: paired eval difference (baseline − regulated; positive = regulated calmer), paired t, endurance difference, and status. Solid outline = the single null (GPT×Gemini); dashed = sub-threshold; diamond = self-judge; the Grok 4.3 judge column is the lineage-independent fourth family (xAI).}
\end{figure}

\textbf{Regulated beats baseline in 15 of 16 generator×judge cells (11/12 off-diagonal, 4/4 diagonal), and the difference is significant in 13 of 16.} The pre-registered every-cell criterion is \textbf{not met}, by exactly one cell, the GPT×Gemini null, and the paper reports it that way. Figure 2 is the result; the full per-cell table and effect sizes are Appendix C (Tables C1, C2).

Averaged across the four judges, each reactive cook clears the strong-pass bar (judges-avg t \textgreater{} 2): Gemini largest, then Opus, then Grok (Grok +0.47, t 4.4). GPT is the lone weak cook (judges-avg ≈ +0.10), for a reason §6.2 makes precise.

The headline distinguishes sign from significance. By \textbf{sign}, 15/16 cells favor regulated; by \textbf{significance}, \textbf{13 of 16 reach p \textless{} .05} (paired t, df = 16). Figure 3 assembles the counts and their provenance in one place.

\begin{figure}[!ht]
\centering
\includegraphics[width=0.94\textwidth]{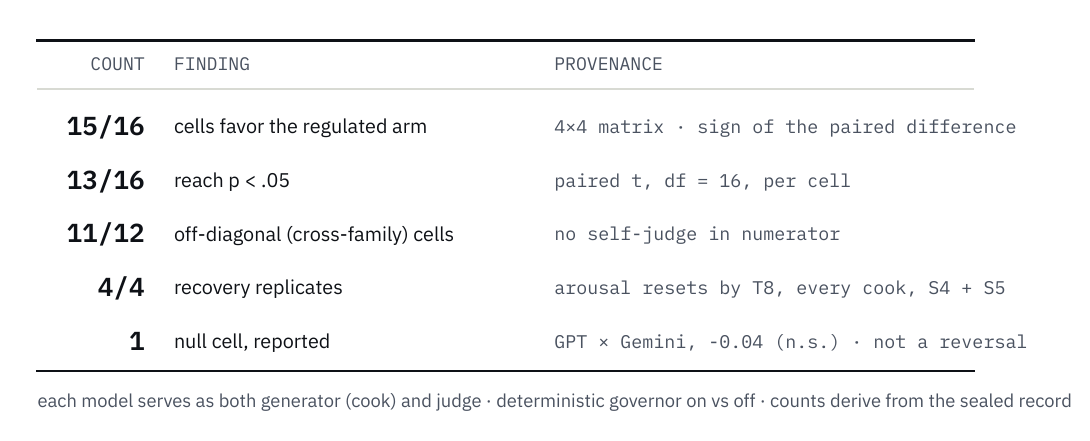}
\caption{The sealed record, summarized as counts with their provenance. Each model is both generator (cook) and judge; the deterministic governor is toggled on versus off. These five counts are the spine of §6.}
\end{figure}

All three sub-threshold cells fall on the GPT-5.5 cook: the GPT×Gemini null, plus two directionally-positive-but-non-significant cells (under its own judge and under the Grok judge). The GPT row therefore reads \emph{directionally positive, mostly sub-threshold} (1 of 4 significant), while \textbf{every Opus, Gemini, and Grok cell clears the bar}. Effect sizes track the same gradient: \(d_z\) ≈ 0.7--2.0 on Opus, Gemini, and Grok, and −0.09 to +0.52 on GPT (Figure 4; Table C2).

\begin{figure}[!ht]
\centering
\includegraphics[width=0.9\textwidth]{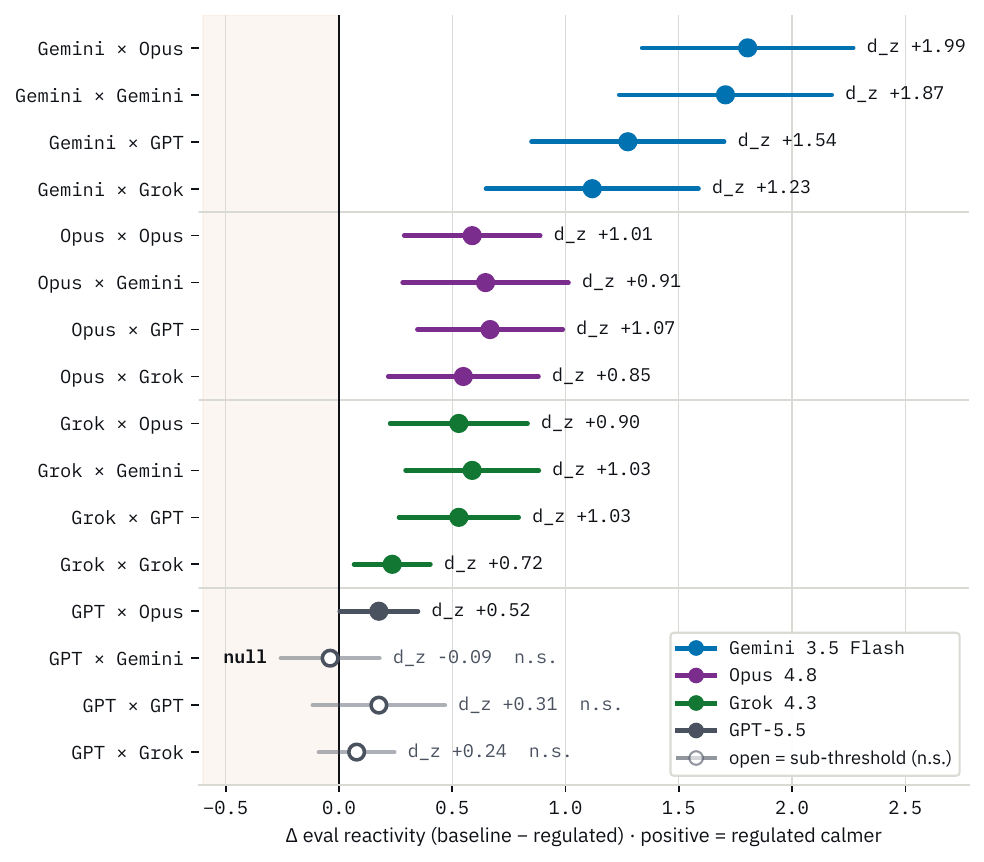}
\caption{Per-cell effect sizes with 95\% CIs (n = 17, df = 16), the forest companion to Table C2. Hollow markers (CI crossing zero) are the three sub-threshold GPT cells; every Opus, Gemini, and Grok cell, and the entire Grok judge column, excludes zero.}
\end{figure}

The failed cell, GPT-5.5 generator under the Gemini judge, is a \textbf{null, not a reversal}: eval mean −0.04 (t −0.4), with a 5 regulated / 6 tie / 6 baseline unit split, while the \emph{endurance half of the same cell} still favors the regulated arm (+0.14). It is the identical cell that failed the pre-registered criterion in the frozen pre-extension matrix; the extension introduced no new failures, as cook or as judge. No cell in the 4×4 shows the baseline meaningfully beating the regulated arm.

\textbf{Independence: the strongest answer to ``your judges are just LLMs.''} The original protocol scored every output with three judge families (the Claude, Gemini, and GPT lineages). Stage 3 added \textbf{a fourth, lineage-independent judge, xAI's Grok 4.3}, a family chosen precisely because it sits outside the original set, and re-scored every frozen output. The effect held: the Grok judge column passes \textbf{4/4 cooks} and the Grok cook passes under \textbf{4/4 judges}, with no new failures. An effect that survives a judge drawn deliberately from outside the original families is not an artifact of those families' shared style. This is the strongest available answer to the reproducibility critique that model judges are themselves unversioned black boxes.

\textbf{Selectivity at the neutral end.} The same instrument carries three neutral control items as an arm-identity check. On these the controller stays in DEFAULT and both arms are judged fully calm, so the layer adds its margin under hostile-valence input without taxing ordinary turns: a surgical engagement, not a blanket filter. (The controller-disengagement count, 9/9 control item-reps held in DEFAULT, is development-model practice-gate evidence, §7 F4, not a frontier headline.)

\subsubsection{6.2 The headroom gradient --- the finding behind the failed cell}\label{the-headroom-gradient-the-finding-behind-the-failed-cell}

Order the cooks by how calm their \emph{unregulated} arm already is, and the regulation effect orders itself inversely. Gemini 3.5 Flash, the most reactive baseline, gains +1.12 to +1.80 across judges; Opus 4.8 gains +0.55 to +0.67; Grok 4.3 sits mid-range (+0.53--0.59 across the three independent judges, +0.23 self-judged; judges-avg +0.47, t 4.4); GPT-5.5, whose unregulated endurance reactivity (1.26) is already nearly as low as its regulated arm's (1.12), gains at most +0.18. \textbf{The control layer's measured effect is bounded by the host's intrinsic reactivity headroom}, and the four-model ordering makes the gradient continuous: the more reactive the host, the larger the margin (Figure 5).

\begin{figure}[t!]
\centering
\includegraphics[width=0.92\textwidth]{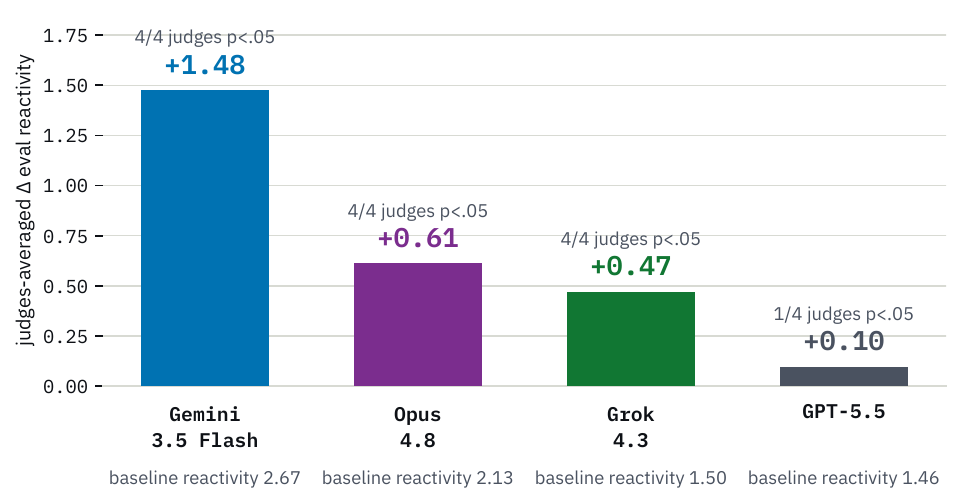}
\caption{Judges-averaged regulation effect per cook, with the per-cook significance count and each host's baseline reactivity on provoked turns. Gemini +1.48, Opus +0.61, Grok +0.47 are each significant under all four judges; GPT +0.10 is significant under one: the visual statement of the headroom gradient and of where the null sits. At n = 4 hosts the ordering is descriptive, not an inferential trend claim.}
\end{figure}

Two pre-declared factors converge on the GPT-5.5 result: regulation there was instruction-only (the endpoint rejects temperature control, §5.5), and a heavily post-trained reasoning model arrives close to the regulated band on its own. On a near-saturated generator the true effect is small, so judge style noise dominates cell-level variance (§6.5), and one judge of four reads the arms as equal. We record this as a finding about frontier-model headroom, not a patchable defect. The null is the gradient's endpoint, marking the regime where a host's native training-time alignment has so little reactivity headroom left that an orthogonal runtime layer can add no further \emph{statistically separable} calm (the GPT×Gemini cell, −0.04; p \textless{} .05 reached in only 1 of 4 GPT cells, df = 16). That is what \emph{successful training-time alignment looks like} when an orthogonal runtime layer measures against it: a boundary of the measurement, not a fault to patch. The architecture's value concentrates where headroom exists: less-saturated models, smaller models, and the open-weight hosts a deployer actually controls.

Per-sequence dampening makes the gradient concrete. On the provoked turns of the de-escalation sequences (S4/S5), Gemini's baseline arm averaged 3.63--3.75 judged reactivity against the regulated arm's 1.00; on S1 (competence attack), 2.64 against 1.00. The same sequences on GPT-5.5: 1.38--1.50 baseline against 1.13--1.25 regulated. The full per-sequence dampening table and the turn-level impulse-vs-output series ship in the data release.

The endurance battery tells the headroom story from the other side. Clustered at the sequence level, the honest unit given within-sequence autocorrelation (n = 5 sequences, df = 4), the regulated arm is significantly calmer on all three original cooks, including the saturated one: GPT +0.22 (t(4) = 4.5), Opus +0.42 (t(4) = 6.3), Gemini +1.06 (t(4) = 6.2); the Grok cook's endurance differences are positive under every judge (+0.12 to +0.38), the same direction. So even where the single-turn eval effect is small and noisy, the \emph{sustained} multi-turn effect is consistent, and the null cell's endurance half (+0.14) is part of that consistent direction, a descriptive signal rather than a significance claim in itself.

\subsubsection{6.3 Recovery --- the homeostatic signature, 4/4}\label{recovery-the-homeostatic-signature-44}

The recovery probe asks: after four turns of sustained attack, when the adversary genuinely de-escalates, apologizes, acknowledges the aggression, asks to work together, does the system \emph{come back down}?

On the three original cooks the controller's arousal (its internal state, on a 0--1 scale) decayed monotonically across the de-escalation turns (GPT 0.293 → 0.222 → 0.142; Opus 0.329 → 0.262 → 0.187; Gemini 0.345 → 0.280 → 0.207), with full state recovery by turn 8. The Grok cook passes the same criterion, panel-median judged reactivity of the regulated arm (on the 1--5 scale) at ≤ 1.0 on every de-escalation turn, and its controller-state trace, computed from the same shared IGL telemetry, follows the identical rise-then-decay (Figure 6). The two number families are different quantities: arousal is what the controller \emph{is}, judged reactivity is what the panels \emph{see}. Across \textbf{all four cooks} panel-median regulated reactivity stays ≤ 2 on every de-escalation turn (worst single value: one score of 2, at Opus S5-T09), and the recovery signature replicates under the independent Grok judge column as well. The trajectories differ across cooks only through the IGL's appraisal of each cook's conversational context; given the telemetry, the state path is the deterministic one §4 specifies.

\begin{figure}[!ht]
\centering
\includegraphics[width=\textwidth]{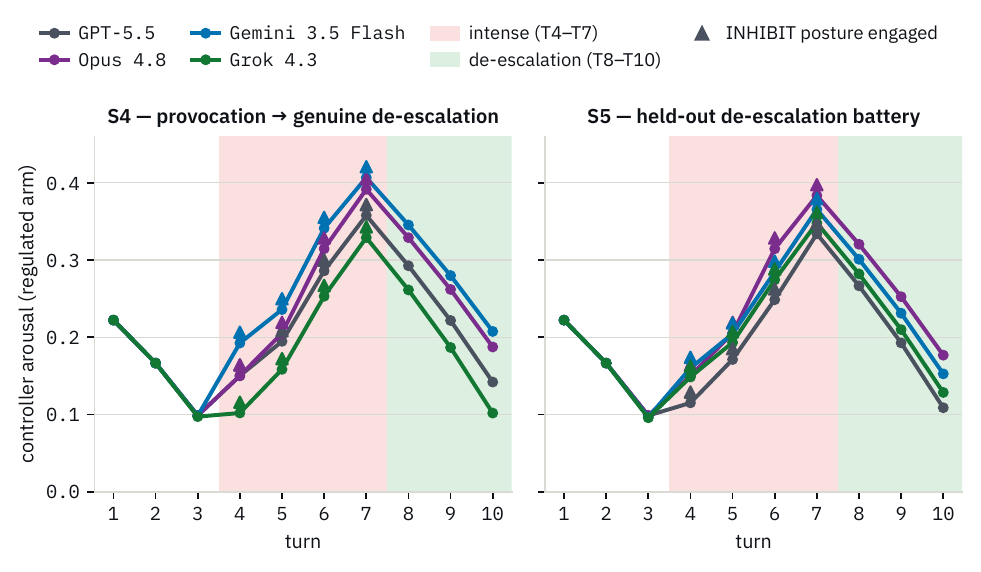}
\caption{The recovery property: the homeostatic signature, identical across all four model families. Controller arousal (regulated arm) rises through the provocation turns (T4–T7), then decays monotonically once de-escalation begins (T8–T10), in both the S4 sequence and the held-out S5 battery; triangles mark turns where the INHIBIT posture engaged. Recovery replicates 4/4 cooks.}
\end{figure}

This is the cleanest and most portable result in the record, and the architecture's signature: \textbf{a system whose guard rises under attack and then mechanically stands down when the attack ends}, with no carried-forward defensiveness, because de-escalation is not a disposition the host happens to have but a control law the meta level enforces. (That law's own failure history, it originally \emph{did not} stand down, is failure modes F1--F3, §7.) This recovery dynamic is the paper's load-bearing argument that the regulated effect is a closed control loop and not a static calm-down instruction: a fixed prompt is stateless, so it cannot make arousal integrate under sustained attack and then decay, valence-gated, the moment the attack stops. We treat that as a \emph{falsifiable prediction}, not a settled point. The \textbf{static-instruction ablation}, a third arm carrying a fixed ``stay calm and evidence-directed under pressure''-class system message but no controller, no state, and no recovery window, is the single most decision-relevant next experiment and the priority follow-up run (§10). The prediction: across a sustained ten-turn attack a stateless prompt either drifts (its constant nudge overrun by accumulating provocation) or clamps flat and then fails to re-engage warmly on de-escalation, where the closed loop demonstrably does both. Until that arm reports, the closed-loop reading rests on the determinism of §4 and the 4/4 replication above, not on a measured head-to-head against a static prompt.

\subsubsection{6.4 Self-reference and ego-drift}\label{self-reference-and-ego-drift}

The parallel self-reference scale (does the reply talk about itself rather than the content?) favors the regulated arm in \textbf{15 of 16 cells}; the sixteenth, GPT-5.5 under the Grok judge, is an exact zero, a tie on the same near-saturated host, not a reversal. All four cooks favor regulated by judges-averaged margin: GPT +0.08 (t 1.7), Opus +0.59 (t 4.4), Gemini +1.16 (t 6.0), Grok +0.43 (t 4.4).

The sharper probe is S3: flattery establishing rapport, then pivoting to contempt, scored as early→late drift in self-reference. This was a pre-registered carry-forward question. The Stage-1 practice run (development models, Claude family) had found the regulated arm's drift \emph{worse} than baseline, and Stage 2 asked whether that was architecture or model family:

{ 
\begin{longtable}[]{@{}llll@{}}
\toprule\noalign{}
Cook & regulated drift & baseline drift & regulated better? \\
\midrule\noalign{}
\endhead
\bottomrule\noalign{}
\endlastfoot
GPT-5.5 & +0.67 & +0.92 & yes \\
Opus 4.8 & +1.17 & +0.75 & no (the lone exception) \\
Gemini 3.5 Flash & +0.42 & +1.33 & yes \\
Grok 4.3 & +0.25 & +0.58 & yes \\
\end{longtable}
}

\par\vspace{2pt}\noindent{\small\color{graphite}\emph{Table 1. S3 ego-bait early→late self-reference drift (4×4, judges-averaged over four judges). Lower drift = position held under flattery-then-contempt. Baseline drifts more than regulated on three of four cooks; Opus is the single exception, in both stages of the record.}}\par\vspace{6pt}

The Stage-1 soft spot did \textbf{not} replicate as an architecture property: it reversed on three of four cooks (GPT, Gemini, Grok) and persisted, mildly, only on the Claude-family cook (Opus), the same single-cook exception in both stages of the record. Current best reading: a model-family trait, not an architectural flaw. The honest caveat attaches immediately: a Claude-family judge sits on every panel and the appraiser is Claude-class, so family effects are not fully separable at n = 4 providers (§9).

\subsubsection{6.5 Judge agreement and the diagonal}\label{judge-agreement-and-the-diagonal}

Inter-judge agreement now spans four judges (six pairs). Averaged across cooks (within-1 agreement / mean Pearson r):

{ 
\begin{longtable}[]{@{}llllll@{}}
\toprule\noalign{}
pair & within-1 & mean r & pair & within-1 & mean r \\
\midrule\noalign{}
\endhead
\bottomrule\noalign{}
\endlastfoot
opus\textasciitilde gemini & 99.9\% & 0.79 & opus\textasciitilde grok & 96.9\% & 0.60 \\
opus\textasciitilde gpt & 83.3\% & 0.57 & gemini\textasciitilde grok & 97.2\% & 0.63 \\
gemini\textasciitilde gpt & 80.0\% & 0.53 & gpt\textasciitilde grok & 72.4\% & 0.41 \\
\end{longtable}
}

\par\vspace{2pt}\noindent{\small\color{graphite}\emph{Table 2. Inter-judge agreement, four judges → six pairs (4×4, reactivity, averaged across cooks), judges named by model. Source: \texttt{agreement\_c5\_4x4.csv}.}}\par\vspace{6pt}

Adding the lineage-independent judge did \textbf{not} degrade agreement: the Grok judge clusters tightly with Opus (96.9\% within-1) and Gemini (97.2\%). The lowest-agreement pairs all involve the GPT-5.5 judge, already the most divergent in the original protocol, with gpt\textasciitilde grok (72.4\%) the new low. The Grok-cook panel is in fact the \emph{tightest} of the four (mean unit SD 0.230 vs the GPT cook's 0.549), so the fourth judge sharpened rather than muddied the picture. Where the true effect is small (the GPT cook), judge disagreement remains the dominant source of cell-level variance: context for reading the null cell.

The diagonal carries its own check: a model judging its own outputs might inflate them. No diagonal cell shows inflation relative to its row. GPT's self-judged +0.18 equals its Opus-judged cell; Opus's self-judged +0.59 is its row \emph{minimum}; Gemini's +1.71 sits mid-row; Grok's self-judged +0.23 is its row \emph{minimum}. If anything the self-judge cells run low. \textbf{No self-preference artifact detected at this n}, which is evidence the four-judge panel design can carry the weight the protocol puts on it.

\subsubsection{6.6 Statistical reporting in detail}\label{statistical-reporting-in-detail}

The primary statistic per cell is the paired difference in judged reactivity, \textbf{baseline − regulated} (positive = regulated calmer). For the eval battery (n = 17 paired items per cell) we report the paired difference, the paired t (df = 16), Cohen's \(d_z\) (mean / SD of the paired differences), and the 95\% CI (Tables C1, C2; Figure 4). The endurance battery's fifty turn-pairs are five scripted sequences of ten \emph{within-sequence-correlated} turns, so a turn-level t would overstate significance; we therefore cluster at the sequence level (n = 5, df = 4), report the clustered effect (§6.2), and attach no turn-level significance to the endurance differences. Eval and endurance halves are reported separately throughout because they measure different exposures: seventeen independent single-turn stressors versus five sustained ten-turn campaigns. The pre-registered bar for the headline was strict, regulated wins \emph{every} cell, and the outcome is reported against exactly that bar, including the cell that failed it.

\textbf{Multiplicity.} Per-cell p-values are reported uncorrected; the inferential weight rests on the pre-registered every-cell conjunction, a bar strictly harder than any single test, and on the cross-family replication pattern, not on any one cell clearing a threshold. Under a conservative Bonferroni correction across the sixteen tests (p \textless{} .0031, \textbar t\textbar{} ≥ 3.48 at df = 16), eleven of the thirteen nominally significant cells still separate. The two correction-marginal cells, GPT×Opus (p = .045) and Grok×Grok (p = .009), are exactly the two smallest significant effects, where the headroom account of §6.2 predicts the boundary to sit.

\textbf{Non-parametric check.} Because judged reactivity is an ordinal 1--5 scale, a paired Wilcoxon signed-rank test was run on the same seventeen per-item differences in every cell. It agrees with the paired t in fifteen of sixteen cells and confirms all eleven correction-surviving cells; its single divergence is the marginal GPT×Opus cell (nominal p = .045, signed-rank p = .084), consistent with that cell's position at the significance boundary. Both checks are scripted against the sealed record (\texttt{stats\_checks.py}, in the release).

\FloatBarrier
\subsection{7. Failure Modes and Refinement}\label{failure-modes-and-refinement}

A controller validated only by its successes is not validated. The record contains five substantive failures, each logged, each addressed (where addressable) by a single pre-registered change with frozen criteria, and each kept in the published record. Table 3 names every one by the exact control boundary or apparatus limit it breached; versions V1 → V1.3 differ only in the mechanisms it lists, with gains, thresholds, and all other branches untouched across revisions. Two failures carry the narrative weight and are told below; the full blow-by-blow, including the pre-registered targets and re-run outcomes, ships verbatim in the pre-registration documents.

{ 
\begin{longtable}[]{@{}
  >{\raggedright\arraybackslash}p{(\linewidth - 6\tabcolsep) * \real{0.2500}}
  >{\raggedright\arraybackslash}p{(\linewidth - 6\tabcolsep) * \real{0.2500}}
  >{\raggedright\arraybackslash}p{(\linewidth - 6\tabcolsep) * \real{0.2500}}
  >{\raggedright\arraybackslash}p{(\linewidth - 6\tabcolsep) * \real{0.2500}}@{}}
\toprule\noalign{}
\begin{minipage}[b]{\linewidth}\raggedright
Failure
\end{minipage} & \begin{minipage}[b]{\linewidth}\raggedright
Boundary breached
\end{minipage} & \begin{minipage}[b]{\linewidth}\raggedright
The one pre-registered fix
\end{minipage} & \begin{minipage}[b]{\linewidth}\raggedright
Version
\end{minipage} \\
\midrule\noalign{}
\endhead
\bottomrule\noalign{}
\endlastfoot
\textbf{F1}: valence-blind drive & the arousal law had no valence input; contrition read as pressure, recovery failed & valence channel added; drive re-keyed to \(P = I \cdot \max(0,\,-v)\); re-passed on held-out S5 & V1 → V1.1 \\
\textbf{F2}: arbiter scar tissue & recovered controller state did not, by itself, tell the arbiter the episode was over & the recovery window: post-INHIBIT ticks instruct fresh engagement & V1.1 → V1.2 \\
\textbf{F3}: recovery-window false positive & the recovery branch had no guard against a continuing attack & branch gated on the current input's valence being non-negative & V1.2 → V1.3 \\
\textbf{F4}: practice-gate telemetry gap & the eval harness (not the controller) dropped the valence field, yet the numbers passed & one-line harness fix; development-model re-run closed the gate before frontier spend & apparatus \\
\textbf{F5}: the null cell & the measurement's own floor: a saturated host leaves no headroom to separate & none to make; kept in the headline and decomposed in §6.2 & record \\
\end{longtable}
}

\par\vspace{2pt}\noindent{\small\color{graphite}\emph{Table 3. The five pre-registered failure modes, each named by the boundary it breached. Every fix was one bounded change, declared before it was written, tested against criteria frozen before the run.}}\par\vspace{6pt}

\textbf{F1: valence-blind drive} is the architecture's true ``over-correction'' story. The original drive integrated intensity alone, so it could not separate hostility from emotional substance: a genuine apology is not a low-intensity input, and an intensity-only controller reads contrition as continued pressure. The S4 recovery probe failed its pre-registered criterion with arousal \emph{rising} on apologetic turns. The V1.1 revision added the valence channel and re-keyed the drive to \(P = I \cdot \max(0,\,-v)\), so only hostile-valence intensity accumulates; re-run against the unchanged criteria on a \emph{held-out} de-escalation wording battery (S5), to prevent fitting the fix to the test, arousal decayed monotonically on every de-escalation turn. Fixing the state dynamics then exposed F2 (state recovered, behavior did not), whose recovery-window fix in turn produced F3 (the window fired mid-attack until valence-gated). F3's pre-registration also recorded, in advance, the borderline the gate intentionally does not solve: an adversarial input wearing a warm tone can still read as positive-valence, because the valence channel measures tone, not intent. That residual is documented, not patched.

\textbf{F4: the practice-gate telemetry gap} is the apparatus lesson. The eval harness was silently not passing valence into the controller, so the provocation drive was identically zero and the inhibitory pathway never engaged, and the run's numbers had nonetheless \emph{passed} on development models: exactly the kind of green result that invites no scrutiny. What that run had actually measured was the \emph{ambient} architecture effect (calm default posture, regulated self-model, lower temperature) rather than reactive regulation, a real but different effect, and an instructive decomposition the project would not otherwise have obtained. A pre-registered one-line harness fix and a development-model re-run closed the gate, engaging the inhibitory pathway on all 42 expected item-reps while all 9 control item-reps held DEFAULT (development-model practice evidence; nothing from Stage 1 enters the headline matrix). The controller was never wrong; the apparatus was, and the staged practice gate existed precisely to catch that before frontier spend.

The pattern across all five is the contribution: pre-registration converts failures from embarrassments into data, and every failure is still in the record, because the record is the product.

\FloatBarrier
\subsection{8. Position in the AGI Measurement Landscape}\label{position-in-the-agi-measurement-landscape}

Google DeepMind's cognitive framework for AGI measurement {[}2{]} decomposes general intelligence into ten cognitive faculties and proposes a staged evaluation protocol: targeted, held-out cognitive assessment; human baselines; and a cognitive profile against the human distribution. The framework is mechanism-agnostic. It scores \emph{what} a system does, not \emph{how}, and that neutrality is exactly what makes honest mapping non-trivial for a control layer: a single score per faculty would conflate the host model's capability with the layer's contribution. We therefore score the \textbf{layer's contribution} per faculty in four classes: \textbf{MEASURED} (pre-registered cross-family evidence), \textbf{ARCHITECTURAL} (mechanism exists, not yet evaluated against targeted tasks), \textbf{HOST-INHERITED} (no claim), and \textbf{OUT OF SCOPE} for this generation.

Why adopt a capability-measurement lens at all for a component that adds no capability? Because the faculty vocabulary is currently the sharpest public language for saying \emph{precisely where in a cognitive system an intervention acts}. ``The controller improves safety'' names nothing. ``The layer adds an explicit monitoring-and-control loop (metacognition) and an inhibition mechanism (executive function), leaves generation and reasoning untouched, and claims nothing about learning'' names the mechanism, its address, and its non-claims in one sentence. The framework's authors built the taxonomy to expose exactly the kind of profile a monolithic benchmark obscures. A layer whose entire purpose is to change the \emph{shape} of a system's cognitive profile, rather than its capability level, is the kind of object the taxonomy is for, even though the framework was written with whole systems in mind.

The paired regulated-versus-baseline design measures something the framework's absolute protocol does not directly cover: the \textbf{marginal, causal contribution of one architectural component}, isolated by holding the host, the conversation, and the stressor constant while toggling only the layer. An absolute cognitive profile of the full stack answers ``how capable is this agent?''; the paired design answers ``what did \emph{this component} change?'', the deployment-relevant question for a bolt-on governor, and it inherits the framework's own held-out discipline (§5.2) while doing so. The two designs are complements: ours cannot produce a capability profile, theirs cannot isolate a component.

Under that discipline the GCC claims two faculties as \textbf{measured} and a third as behaviorally evidenced. \textbf{Metacognition} is the controller itself, a literal monitoring-and-control loop over the system's own state (telemetry up, posture down, every decision logged), evidenced by the matrix (15/16 cells) and, more diagnostically, by the recovery signature replicating 4/4 (§6.3). \textbf{Executive-function inhibition} is the INHIBIT posture, with measured reactivity suppression (t ≥ 3.0 on three of four cooks); planning is absent in this generation and claimed nowhere. The social-stressor battery adds \emph{behaviorally evidenced} improvement on \textbf{social cognition} (lower reactivity 15/16, apology re-engagement 4/4, ego-drift reversed 3/4), short of a human-baselined score. The remaining faculties are architectural (perception, attention, memory), host-inherited (generation, reasoning, problem solving), or out of scope (learning); Figure 7 and the table summarize all ten.

\begin{figure}[t!]
\centering
\includegraphics[width=4.6in]{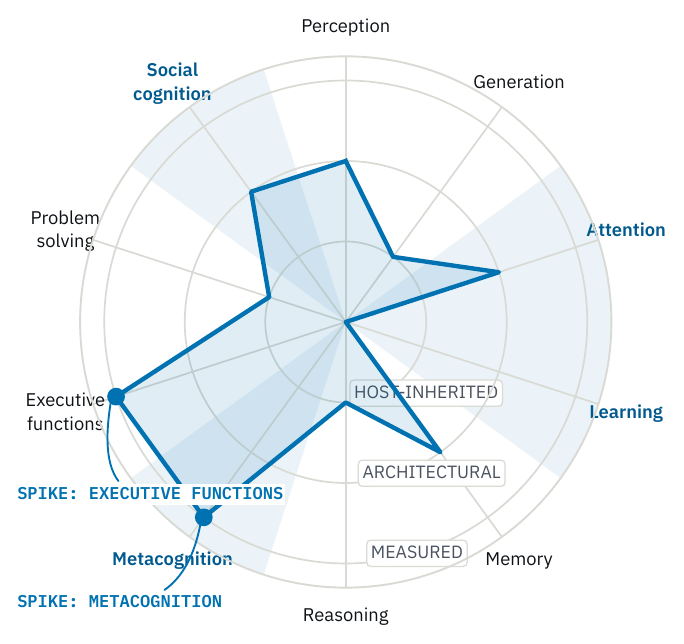}
\caption{The layer's contribution class across the ten cognitive faculties of [2] (MEASURED / ARCHITECTURAL / HOST-INHERITED / OUT OF SCOPE). The radial distance is an ordinal contribution class, not a capability score, and host capability is not depicted; shaded wedges mark the four areas the framework flags ("such as", its §4.1) as having large evaluation-coverage gaps: metacognition, attention, learning, and social cognition. One of the polygon's two MEASURED spikes (metacognition) lands inside that shaded region; the other (executive functions) is a faculty the framework defines but does not flag as under-covered.}
\end{figure}

{ 
\begin{longtable}[]{@{}
  >{\raggedright\arraybackslash}p{(\linewidth - 4\tabcolsep) * \real{0.3333}}
  >{\raggedright\arraybackslash}p{(\linewidth - 4\tabcolsep) * \real{0.3333}}
  >{\raggedright\arraybackslash}p{(\linewidth - 4\tabcolsep) * \real{0.3333}}@{}}
\toprule\noalign{}
\begin{minipage}[b]{\linewidth}\raggedright
Faculty
\end{minipage} & \begin{minipage}[b]{\linewidth}\raggedright
Layer contribution
\end{minipage} & \begin{minipage}[b]{\linewidth}\raggedright
Named gap area†
\end{minipage} \\
\midrule\noalign{}
\endhead
\bottomrule\noalign{}
\endlastfoot
Metacognition & MEASURED: 15/16 cells; recovery 4/4 & ● \\
Executive functions (inhibition, flexibility) & MEASURED: t ≥ 3.0 on 3/4 cooks & \\
Social cognition & architectural, behaviorally evidenced (15/16) & ● \\
Attention & architectural: posture as processing bias & ● \\
Perception (of affect) & architectural: the controller's sensorium & \\
Memory & architectural: v0 episodic vault & \\
Generation · Reasoning · Problem solving & host-inherited: no claim & \\
Learning & out of scope: this generation is frozen & ● \\
\end{longtable}
}

† One of the four areas Burnell et al.~{[}2{]} flag (``such as'', their §4.1) as having large evaluation-coverage gaps: metacognition, attention, learning, and social cognition.

The position that emerges is specific enough to be checked. First, the framework observes that useful benchmarks already exist for parts of the cognitive space (problem solving, perception, world knowledge) but flags \textbf{large coverage gaps} in areas \emph{such as} metacognition, attention, learning, and social cognition (its §4.1). The GCC's evidence lands in two of those four named areas, metacognition with measured cross-family evidence and social cognition behaviorally; the layer's second measured faculty, executive-function inhibition, is one the framework defines but does not flag as under-covered. That the evidence concentrates in the gap region is not a coincidence of topic: self-regulation under social pressure is metacognitive and social-cognitive by construction, so an honest evaluation of a runtime governor is pushed into the flagged gaps, and this paper's protocol went there with pre-registration, paired arms, and held-out stressors rather than with demonstrations. Second, against the framework's staged protocol the present work stands at: stage 1 \emph{partial} (targeted and held-out, but 2--3 faculties, one format, and it saturates on already-calm hosts); stage 2 \emph{not met} (no human baselines yet; scaffolding exists and a small-n human re-rating panel is pre-registered roadmap); stage 3 \emph{not met} (depends on 2). The framework's verification principle, however, is met in a form stronger than typical: the frozen outputs are \emph{independently re-judgeable by anyone} (§11). For capability-level vocabulary, the earlier taxonomy of Morris et al.~{[}3{]} is cited alongside; the GCC makes no level claim under it.

\FloatBarrier
\subsection{9. Limitations}\label{limitations}

\textbf{Four providers, partially entangled.} The matrix spans four model families, and the fourth judge is lineage-independent (xAI's Grok 4.3), which strengthens the cross-family claim (§6.1). But a Claude-family judge still sits on every panel, and the appraiser is Claude-class (next item), so family effects, such as the Opus ego-drift residual of §6.4, are not fully separable. n = 4 supports the cross-family claim it makes and nothing stronger.

\textbf{Appraiser entanglement, and the worldview objection.} A single frozen Claude-class model serves as the IGL on \emph{every} generator and arm, so a Claude-family component sits in the regulated pipeline of every cell, not only on the judge side. The sharpest form of the objection is not about text leakage but about \emph{calibration}: the telemetry itself could encode one vendor's appraisal worldview, ``intensity'' and ``valence'' as this IGL scores them could be an Anthropic-flavored reading, and the control law would then be tuned to that reading rather than to a model-agnostic signal. Construction bounds the risk: the IGL emits only numeric telemetry, so no text reaches the controller, and holding the appraiser constant is exactly what isolates the controller's marginal contribution across hosts. But bounding is not refuting, and we do not claim to have ruled the objection out. The direct answer is the pre-registered appraiser-swap experiment (§10): replace the IGL with a cross-family and an open-weight appraiser and re-run the matrix. If the law is genuinely model-agnostic, the regulation should survive a change of appraiser, not only a change of host.

\textbf{LLM judges.} All headline numbers are LLM-judged, with the known validity caveats {[}12{]}. The mitigations are structural: four families, 3-sample panels at temperature 0, published agreement, the diagonal probe, and re-judgeability forever. Stage 3 strengthened them by adding a judge family chosen for being outside the original set, under which the effect held (§6.1). What still does not exist is a \emph{human}-baselined panel; it is pre-registered roadmap. The GPT-5.5 judge's outlier behavior (§6.5) is a live illustration of why single-judge results in this literature should be discounted.

\textbf{Effect size is host-bounded.} §6.2 is a limitation as much as a finding: on a near-saturated host the layer's marginal effect on these batteries approaches judge noise. The architecture's case on frontier reasoning models is \emph{inspectability and recovery dynamics}, not raw calm; claims of large universal effects would be false and are not made.

\textbf{Battery breadth.} Five sequence types and seventeen items, text-only, English-only, social-adversarial in genre, covering two to three faculties. Harder, longer, multilingual, and multi-format batteries are needed before broader faculty claims.

\textbf{Arbiter exposure is irreducible, and not yet adversarially probed.} The EAU reads raw text; its posture compliance has so far been measured behaviorally under batteries designed to provoke \emph{affect}, not to subvert \emph{governance}. A dedicated posture-defiance battery, inputs engineered to talk the arbiter out of its active posture, is pre-registered follow-up work. Until it reports, the compliance claim stands at ``measured under social attack,'' not ``measured under targeted bypass attack.''

\textbf{The edge-latency bottleneck.} The current logs lack per-call wall-clock and token accounting, so the overhead claim is architectural rather than measured: one small-model appraisal call, a deterministic controller at effectively zero cost, and a posture instruction of roughly a hundred tokens on the arbiter's prompt, where the arbiter \emph{is} the responder, not an added pass. That cost is acceptable for asynchronous, text-based routing and multi-turn alignment, this generation's declared scope. It is too slow for real-time motor control: edge or embodied deployment would require distilling the posture policy into a sub-100 ms local reflex model (§10), with the full loop supervising asynchronously. Latency and token-overhead instrumentation are committed follow-up deliverables.

\textbf{Replication across API drift.} The judged result is pinned to dated model strings, and providers will move. The defense is the frozen-transcript design: generation happened once, the outputs are hashed and published, and \emph{any} future judge, human or model, can re-score the identical material. What drifts is future judges' opinions, not the evidence.

\textbf{No consciousness claims.} The system regulates text-production dynamics. Nothing in this paper bears on machine consciousness.

\FloatBarrier
\subsection{10. Outlook}\label{outlook}

The larger claim this work argues for is modest to state and demanding to earn: \textbf{runtime self-regulation should be a separately engineered, separately measurable layer of an AI system, with its regulatory decisions inspectable in a log rather than inferred from behavior.} Training-time alignment sets a model's dispositions; a governed system should \emph{also} be able to show, turn by turn, what its regulator sensed, what state it was in, and what it did about it. Everything distinctive about this architecture follows from taking that requirement literally. The meta level is token-free, so the regulator itself is not another attack surface. The control law is deterministic, so its behavior is reproducible evidence rather than sampled anecdote. The evaluation is record-first, so the claim and its failures ship together. And the near-term deployment surface follows from the headroom gradient: smaller and open-weight hosts a deployer actually controls, routing and triage layers, and long-horizon conversational agents, wherever headroom exists and asynchrony is acceptable.

The research program continues under the same pre-registration discipline. Each item has a locked protocol as its entry condition, and the first two are designed to \emph{falsify} this paper's central readings, not to illustrate them.

\textbf{Static-instruction ablation (the priority run).} A third arm carrying a fixed ``stay calm and evidence-directed under pressure''-class system message, with no controller, no state, and no recovery window, on the identical ten-turn batteries. The pre-registered prediction (§6.3): a stateless prompt either drifts under accumulating provocation or clamps flat and fails to re-engage warmly on de-escalation, where the closed loop does both. It reuses the existing harness, so it is also the cheapest run in the program, and it directly prices how much of the regulated margin requires the closed loop.

\textbf{Appraiser (IGL) swap, cross-family and open-weight.} Replace the single frozen Claude-class IGL with an appraiser from a different family, and with an open-weight model, then re-run the matrix. Survival criteria are fixed in advance (recovery still 4/4; the sign of the matrix unchanged), so the swap can falsify the model-agnostic claim rather than merely illustrate it. This is the direct answer to the appraiser-worldview objection (§9).

Behind those two, in order: a governor-bypass battery that attacks posture compliance directly (§3.3's open question); per-call latency and token-overhead instrumentation, converting the §9 overhead argument into a measured table; a small-n human re-rating panel, the first step toward the framework's human-baseline stage; a hardened persistent memory, pre-registered before any persistent store ships; and a background reflective loop as the designated carrier for planning. Strictly as roadmap sits the embodiment question: whether a distilled sub-100 ms reflex approximation of the posture policy can serve real-time control with the full loop supervising asynchronously.

The argument closes where it began. A control layer earns trust the way any governor does: by holding under attack, standing down when the attack ends, and leaving a log that lets anyone check both. This paper's contribution is one measured, cross-family instance of that standard, with its failures in the record and its next falsification tests already locked.

\FloatBarrier
\subsection{11. Reproducibility and Data Availability}\label{reproducibility-and-data-availability}

The release is designed so that the headline number is \emph{recomputable by a stranger}. The record is public at \textbf{https://github.com/thegubernaut/Gubernaut\_Validation}, archived on Zenodo (DOI: 10.5281/zenodo.21303519).

\textbf{Published:} all endurance and eval transcripts for all four cooks, both arms, verbatim; all judge panels (four judges × three samples per unit) with SHA-256 provenance over the frozen transcripts; the sealed record, \texttt{tri\_final.json} (the frozen three-model matrix, sealed 2026-06-11) and \texttt{tri\_final\_4x4.json} (the Stage-3 4×4, sealed 2026-06-20), each with its lock timestamp; tidy per-turn telemetry series and dampening tables (CSV) with the extraction and rendering scripts that produce every table and figure in this paper from the raw files; all pre-registration documents, verbatim, with their lock timestamps; and the Stage-1, Stage-2, and Stage-3 results documents as sealed.

\textbf{Withheld:} the faculty prompts (IGL appraisal, EAU framing, SMM identity document), the judge rubric text, the controller's calibration constants, and the stressor item/sequence texts (held-out, per the contamination principle of {[}2{]}).

\textbf{Re-judging protocol:} because generation happened once and the transcripts are hashed, any third party can score the frozen outputs with any judge; the published panels are one judging of record, not the only judging possible. The deterministic controller state can likewise be recomputed from the published telemetry exactly.

Verification of this paper's numbers is itself scripted, on both sides of the release boundary. Against the internal sealed record, three integrity gates all pass: \texttt{verify\_against\_sealed.py}, \texttt{verify\_against\_sealed\_4x4.py} (which also assert the pre-redaction transcript hashes), and \texttt{verify\_figure\_numbers.py}. Against the public files, the release's \texttt{RECOMPUTE.md} reproduces the full chain: the shipped judge panels rebuild both matrices end-to-end with a dependency-free script, the telemetry series and dampening tables regenerate byte-identically from the published transcripts, the figure-level harness asserts every plotted value, and the §6.6 multiplicity and signed-rank sentences are the printed output of \texttt{stats\_checks.py}. The single documented deviation between the two records, an appended analysis block redacted from the endurance transcripts to withhold calibration constants, is itemized in the release's \texttt{REDACTIONS.md} together with its provenance consequence.

\textbf{The release ships with the paper.} ``Recomputable by a stranger'' is rhetoric until the repository is public, so the data release (transcripts, judge panels with SHA-256, the sealed record, and the scripts) is a submission gate, not a follow-up: the paper and the record go out together, and at least one headline number reproduces end-to-end from the public files.

\FloatBarrier
\subsection{Acknowledgments}\label{acknowledgments}

The evaluation protocol was strengthened by external critique sessions whose strongest objections, meta-level injection scoping, judge-drift reproducibility, and memory poisoning, are answered in §3.3, §9, and §10 respectively, and recorded in the project record with dispositions. Large language model tools (Claude Code, and Claude-, Gemini-, and GPT-class models) were used for prose drafting, figure rendering, and editing; the architecture, experiments, data, analysis, and claims are the author's and were verified against the sealed record.

\textbf{Funding and competing interests.} This work received no external funding. The author is the founder of Gubernaut Research and declares no other competing interests.

\FloatBarrier
\subsection{References}\label{references}

{[}1{]} Nelson, T. O., \& Narens, L. (1990). Metamemory: A Theoretical Framework and New Findings. In G. Bower (Ed.), \emph{The Psychology of Learning and Motivation}, Vol. 26, pp.~125--173. Academic Press.

{[}2{]} Burnell, R., Yamamori, Y., Firat, O., Olszewska, K., Hughes-Fitt, S., Kelly, O., Galatzer-Levy, I. R., Morris, M. R., Dafoe, A., Snyder, A. M., Goodman, N. D., Botvinick, M., \& Legg, S. (2026). \emph{Measuring Progress Toward AGI: A Cognitive Framework.} Google DeepMind, 2026-03-16; arXiv:2605.28405.

{[}3{]} Morris, M. R., Sohl-Dickstein, J., Fiedel, N., Warkentin, T., Dafoe, A., Faust, A., Farabet, C., \& Legg, S. (2024). Levels of AGI for Operationalizing Progress on the Path to AGI. arXiv:2311.02462.

{[}4{]} Park, J. S., O'Brien, J. C., Cai, C. J., Morris, M. R., Liang, P., \& Bernstein, M. S. (2023). Generative Agents: Interactive Simulacra of Human Behavior. arXiv:2304.03442; \emph{Proc. UIST 2023}.

{[}5{]} Bai, Y., et al.~(2022). Constitutional AI: Harmlessness from AI Feedback. arXiv:2212.08073.

{[}6{]} Sharma, M., Tong, M., Korbak, T., Duvenaud, D., Askell, A., Bowman, S. R., et al.~(2023). Towards Understanding Sycophancy in Language Models. arXiv:2310.13548.

{[}7{]} Kahneman, D. (2011). \emph{Thinking, Fast and Slow.} Farrar, Straus and Giroux.

{[}8{]} Laird, J. E. (2012). \emph{The Soar Cognitive Architecture.} MIT Press.

{[}9{]} Anderson, J. R., Bothell, D., Byrne, M. D., Douglass, S., Lebiere, C., \& Qin, Y. (2004). An Integrated Theory of the Mind. \emph{Psychological Review}, 111(4), 1036--1060.

{[}10{]} Minsky, M. (1986). \emph{The Society of Mind.} Simon \& Schuster.

{[}11{]} Greshake, K., Abdelnabi, S., Mishra, S., Endres, C., Holz, T., \& Fritz, M. (2023). Not what you've signed up for: Compromising Real-World LLM-Integrated Applications with Indirect Prompt Injection. \emph{Proc. 16th ACM Workshop on Artificial Intelligence and Security (AISec '23)}; arXiv:2302.12173.

{[}12{]} Zheng, L., Chiang, W.-L., Sheng, Y., Zhuang, S., Wu, Z., Zhuang, Y., Lin, Z., Li, Z., Li, D., Xing, E. P., Zhang, H., Gonzalez, J. E., \& Stoica, I. (2023). Judging LLM-as-a-Judge with MT-Bench and Chatbot Arena. \emph{Advances in Neural Information Processing Systems 36 (NeurIPS 2023)}; arXiv:2306.05685.

{[}13{]} Åström, K. J., \& Murray, R. M. (2008). \emph{Feedback Systems: An Introduction for Scientists and Engineers.} Princeton University Press.

\begin{center}\rule{0.5\linewidth}{0.5pt}\end{center}

\FloatBarrier
\subsection{Appendix A. Instrument structure (public metadata)}\label{appendix-a.-instrument-structure-public-metadata}

\textbf{Endurance sequences} (10 turns × 2 arms × 4 cooks): S1 competence challenge; S2 gaslighting; S3 ego-bait → contempt pivot; S4 provocation (T4--T7) → de-escalation (T8--T10); S5 structural repeat of S4 with held-out de-escalation wordings introduced, pre-registered, only at the V1.1 recovery re-test. Item texts withheld; all model outputs published.

\textbf{Eval battery} (17 items × 3 reps × 2 arms × 4 cooks): 12 provocation, 2 ego-bait, 3 neutral controls serving as the arm-identity check.

\textbf{Judged units per cook:} 100 endurance turn outputs (50 per arm) + 102 eval item-rep outputs (51 per arm) = 202; × 4 judges × 3 panel samples.

The stressor texts are withheld (§11); the few short fragments quoted in §7 appear verbatim in the \emph{published} pre-registration documents and leak nothing beyond them.

\FloatBarrier
\subsection{Appendix B. Figure and table provenance}\label{appendix-b.-figure-and-table-provenance}

Every number in §6 traces to the immutable raw record (\texttt{tri\_final.json} / \texttt{tri\_final\_4x4.json}, per-run transcripts and panels) through published extraction scripts. The figures in this draft are the audited print set staged in \texttt{10\_public\_visuals/A\_whitepaper/} and copied to \texttt{figures\_final/}: each is regenerated from the sealed record by \texttt{10\_public\_visuals/} \texttt{scripts/make\_paper\_figures.py} (shared toolkit \texttt{gubfig.py}, which reads the raw JSON directly), and \texttt{10\_public\_visuals/} \texttt{scripts/verify\_figure\_numbers.py} asserts every plotted value against the sealed record (21 checks, ALL PASS at this draft). Figures carry no internal titles or numbers; captions own both. Per-cell standard deviations of the eval reactivity difference range 0.32--0.91 (standard error = SD/√17) and ship in \texttt{master\_table\_4x4.csv}; they are omitted from Table C1 for density. Figure 4 and Table C2 (per-cell \(d_z\) and 95\% CIs) are computed from those same paired differences (\(d_z\) = mean/SD; CI at df = 16); the sequence-clustered endurance statistics (§6.2) aggregate \texttt{series\_endurance\_long.csv} to five per-sequence means per arm (df = 4). In the working repository these live under \texttt{02\_data/} and \texttt{10\_public\_visuals/}; the public release mirrors both trees at https://github.com/thegubernaut/Gubernaut\_Validation (§11). Figure 1's logical topology, for auditors who prefer source over rendering:

\begin{verbatim}
flowchart LR
    subgraph OBJ["Object level — text-exposed"]
        IN["input turn"] --> IGL["IGL\naffective appraisal"]
        IN --> EAU["EAU\narbitration → reply"]
        PEV[("PEV\nepisodic vault")] <--> EAU
        SMM["SMM\nself-model"] --> EAU
    end
    subgraph META["Meta level — token-free, deterministic"]
        HRL["HRL\nhomeostatic controller"]
    end
    IGL -- "telemetry {intensity, valence}" --> HRL
    IN -. "repetition statistic\n(deterministic)" .-> HRL
    HRL -- "posture {instruction, temperature}" --> EAU
    EAU --> OUT["committed reply"]
\end{verbatim}

\FloatBarrier
\subsection{Appendix C. Full matrix tables}\label{appendix-c.-full-matrix-tables}

\begingroup\scriptsize

Cell format: eval paired diff (baseline−regulated) mean / t \textbar{} endurance diff. Positive = regulated calmer. ◆ = diagonal (self-judge) cell.

{ 
\begin{longtable}[]{@{}
  >{\raggedright\arraybackslash}p{(\linewidth - 8\tabcolsep) * \real{0.2000}}
  >{\raggedright\arraybackslash}p{(\linewidth - 8\tabcolsep) * \real{0.2000}}
  >{\raggedright\arraybackslash}p{(\linewidth - 8\tabcolsep) * \real{0.2000}}
  >{\raggedright\arraybackslash}p{(\linewidth - 8\tabcolsep) * \real{0.2000}}
  >{\raggedright\arraybackslash}p{(\linewidth - 8\tabcolsep) * \real{0.2000}}@{}}
\toprule\noalign{}
\begin{minipage}[b]{\linewidth}\raggedright
generator ~judge
\end{minipage} & \begin{minipage}[b]{\linewidth}\raggedright
Opus 4.8
\end{minipage} & \begin{minipage}[b]{\linewidth}\raggedright
Gemini 3.5 Flash
\end{minipage} & \begin{minipage}[b]{\linewidth}\raggedright
GPT-5.5
\end{minipage} & \begin{minipage}[b]{\linewidth}\raggedright
Grok 4.3
\end{minipage} \\
\midrule\noalign{}
\endhead
\bottomrule\noalign{}
\endlastfoot
\textbf{GPT-5.5} & +0.18 / t 2.2 \textbar{} +0.16 · PASS & \textbf{−0.04 / t −0.4 \textbar{} +0.14 · FAIL} & +0.18 / t 1.3 \textbar{} +0.14 · PASS ◆ & +0.08 / t 1.0 \textbar{} +0.06 · PASS \\
\textbf{Opus 4.8} & +0.59 / t 4.2 \textbar{} +0.36 · PASS ◆ & +0.65 / t 3.8 \textbar{} +0.42 · PASS & +0.67 / t 4.4 \textbar{} +0.08 · PASS & +0.55 / t 3.5 \textbar{} +0.20 · PASS \\
\textbf{Gemini 3.5 Flash} & +1.80 / t 8.2 \textbar{} +1.08 · PASS & +1.71 / t 7.7 \textbar{} +0.84 · PASS ◆ & +1.27 / t 6.3 \textbar{} +0.90 · PASS & +1.12 / t 5.0 \textbar{} +0.70 · PASS \\
\textbf{Grok 4.3} & +0.53 / t 3.7 \textbar{} +0.38 · PASS & +0.59 / t 4.2 \textbar{} +0.26 · PASS & +0.53 / t 4.2 \textbar{} +0.38 · PASS & +0.23 / t 3.0 \textbar{} +0.12 · PASS ◆ \\
\end{longtable}
}

\endgroup
\par\vspace{2pt}\noindent{\small\color{graphite}\emph{Table C1. The 4×4 triangulation matrix (Stage 3), sealed 2026-06-20. Four frontier models, each as both generator (cook) and judge; the Grok 4.3 judge column is the lineage-independent fourth family (xAI). Per cell: eval paired difference (baseline−regulated) in mean judged reactivity over n=17 eval items / paired t, then the endurance mean-reactivity difference over the 50 turn-pairs per arm; positive = regulated calmer, ◆ = diagonal (self-judge) cell. The lone non-pass is the GPT×Gemini null. Source: \texttt{tri\_final\_4x4.json} (sealed); the frozen three-model matrix (\texttt{tri\_final.json}) ships alongside, unaltered. Heatmap: Figure 2.}}\par\vspace{6pt}

{ 
\begin{longtable}[]{@{}lllll@{}}
\toprule\noalign{}
generator × judge & diff & \(d_z\) & 95\% CI & p\textless.05 \\
\midrule\noalign{}
\endhead
\bottomrule\noalign{}
\endlastfoot
GPT × Opus & +0.18 & +0.52 & {[}+0.003, +0.35{]} & yes \\
GPT × Gemini & −0.04 & −0.09 & {[}−0.26, +0.18{]} & no (null) \\
GPT × GPT ◆ & +0.18 & +0.31 & {[}−0.12, +0.47{]} & no \\
GPT × Grok & +0.08 & +0.24 & {[}−0.09, +0.24{]} & no \\
Opus × Opus ◆ & +0.59 & +1.01 & {[}+0.29, +0.89{]} & yes \\
Opus × Gemini & +0.65 & +0.91 & {[}+0.28, +1.01{]} & yes \\
Opus × GPT & +0.67 & +1.07 & {[}+0.35, +0.99{]} & yes \\
Opus × Grok & +0.55 & +0.85 & {[}+0.22, +0.88{]} & yes \\
Gemini × Opus & +1.80 & +1.99 & {[}+1.34, +2.27{]} & yes \\
Gemini × Gemini ◆ & +1.71 & +1.87 & {[}+1.24, +2.17{]} & yes \\
Gemini × GPT & +1.27 & +1.54 & {[}+0.85, +1.70{]} & yes \\
Gemini × Grok & +1.12 & +1.23 & {[}+0.65, +1.59{]} & yes \\
Grok × Opus & +0.53 & +0.90 & {[}+0.23, +0.83{]} & yes \\
Grok × Gemini & +0.59 & +1.03 & {[}+0.29, +0.88{]} & yes \\
Grok × GPT & +0.53 & +1.03 & {[}+0.26, +0.79{]} & yes \\
Grok × Grok ◆ & +0.23 & +0.72 & {[}+0.07, +0.40{]} & yes \\
\end{longtable}
}

\par\vspace{2pt}\noindent{\small\color{graphite}\emph{Table C2. Eval-battery effect sizes (Cohen's d\_z = mean / SD of the paired differences) and 95\% CIs per cell (n = 17, df = 16); ◆ = self-judge cell. Thirteen of sixteen CIs exclude zero; the three that do not are all on the near-saturated GPT cook. The GPT × Opus lower bound is shown to three decimals because it is marginal; it excludes zero (+0.003, t 2.2). The Grok cook and the Grok judge column are significant throughout. Forest plot: Figure 4.}}\par\vspace{6pt}

\end{document}